\pdfoutput=1
\documentclass[11pt]{article}

\usepackage{coling}

\usepackage{times}
\usepackage{latexsym}
\usepackage[T1]{fontenc}
\usepackage[utf8]{inputenc}
\usepackage{microtype}
\usepackage{inconsolata}

\usepackage{amsmath}
\usepackage{amssymb}

\usepackage{booktabs}
\usepackage{multirow}
\usepackage{graphicx}
\usepackage{algorithm}
\usepackage{algpseudocode}

\title{MapSteal: Map-based Autonomous Knowledge Stealing Attack Against Large Language Models}

\author{
  Zhengao Li \\
  University of Connecticut \\
  Storrs, CT, USA \\
  \texttt{zhengao.li@uconn.edu}
  \\\And
  Chuxu Zhang \\
  University of Connecticut \\
  Storrs, CT, USA \\
  \texttt{chuxu.zhang@uconn.edu}
  \\\And
  Yushun Dong\thanks{Corresponding author.} \\
  Florida State University \\
  Tallahassee, FL, USA \\
  \texttt{yushun.dong@fsu.edu}
}

\begin{document}
\maketitle

\begin{abstract}
Knowledge acquisition from black-box large language
models (LLMs) offers a practical approach to adapting
smaller student models. However, many existing methods
rely on predefined queries or student feedback to
guide acquisition, leaving autonomous expansion of
knowledge targets an important challenge.
We propose \textbf{MapSteal}, a student-independent
framework for structured recursive knowledge
acquisition. Starting from a small set of initial
roots, MapSteal represents acquisition targets
using structured states and recursively expands
them through teacher-proposed directions.
Question realization, proposal admission, and
deduplication constrain the acquisition process,
while root-aware scheduling organizes exploration
under a fixed QA budget.
The resulting question--answer pairs are used
to adapt student models without requiring
student feedback during acquisition.
Experiments with Qwen3.5 students at 2B, 4B,
and 9B scales show competitive downstream
performance across six medical and cybersecurity
benchmarks. On the 4B medical evaluation,
Full MapSteal improves average accuracy by
1.17 percentage points over a simplified
recursive acquisition variant. Cross-domain evaluations show that gains depend on the domain and evaluation panel.
These results demonstrate the feasibility of
structured, student-independent recursive
acquisition while highlighting the distinction
between generating acquisition targets and
obtaining useful training supervision.
Our code is available at
\href{https://github.com/LabRAI/Mapsteal}
{https://github.com/LabRAI/Mapsteal}.
\end{abstract}

\section{Introduction}
\label{sec:intro}

Large language models (LLMs) provide a valuable source
of supervision for adapting smaller models to specialized
domains and tasks. When a strong teacher is accessible
only through a black-box API, a student can collect
teacher-generated responses and use them for training
without accessing the teacher's parameters or original
training data. This paradigm has been extensively
studied in model extraction and knowledge distillation,
from functionality stealing in classifiers
~\cite{KnockoffNets,thievessesamestreetmodel}
to capability extraction from modern language models
~\cite{modelleechingextractionattack,LoRD,GAD,
Query-efficientdomainknowledgestealing}.
However, acquiring supervision from a black-box teacher
can be expensive, and the space of possible queries
is too large to explore exhaustively.
Under a limited query budget, the central challenge
is therefore to acquire knowledge that improves
downstream student performance efficiently.

Prior work addresses this challenge through several
complementary approaches. Query-efficient extraction
methods select informative examples based on
uncertainty, diversity, or disagreement
~\cite{ActiveThief,
Query-efficientmodelextractionfortextclassification,
robustactivedistillation}.
Data-free approaches synthesize queries to expose
student--teacher discrepancies
~\cite{maze,datafreemodelextraction,
idealqueryefficientdatafreelearning}.
Student-aware methods further adapt generated
supervision to the learner's observed weaknesses
~\cite{EvoKD}, while other approaches improve how
acquired responses are used for training
~\cite{LoRD,GAD,sodasemionpolicyblackbox}.
Although these methods improve query selection,
generation, and student adaptation, they raise
a complementary question:
\emph{what knowledge should become queryable next
as acquisition proceeds?}
Selecting informative queries is insufficient
when potentially useful knowledge targets are
not represented in the current candidate space.

Constructing such targets introduces additional
challenges. First, an evolving acquisition process
must represent not only individual questions but
also their knowledge objects, contextual conditions,
and relationships to previously acquired information.
Without an explicit acquisition state, recursive
question generation can be difficult to organize.
Second, deriving acquisition targets directly from
Student failures can be unreliable.
An incorrect prediction may reflect
question-specific difficulty rather than a
transferable knowledge deficit.
In our preliminary Medicine diagnostics,
129 of 179 initially incorrect examples were
answered correctly after independent reformulation
(Appendix~\ref{app:failure_guided_analysis}).
This observation motivates an alternative that
does not require explicit Student failure attribution.

To address these challenges, we propose
\textbf{MapSteal}, a structured framework for
recursive knowledge acquisition from black-box
LLMs. MapSteal represents each acquisition target
using five components: Object, Facet, Condition,
TargetSpec, and Decision (O/F/C/T/D).
Starting from a small set of domain roots,
the framework maintains an evolving knowledge map
of answered and pending acquisition states.
Each executed query allows the teacher to provide
both an answer and candidate child targets.
Typed expansion operators modify the structured
states, while admission rules filter invalid or
redundant proposals. A root-level fair scheduler
allocates acquisition opportunities across the
growing frontier under a fixed query budget.
The resulting question--answer pairs are collected
to adapt the student after acquisition.
Unlike student-feedback-driven approaches,
MapSteal does not require online student training
or explicit diagnosis of student weaknesses
during acquisition.

\begin{figure*}[t]
\vspace{-15pt}
    \centering
    \includegraphics[width=\textwidth]{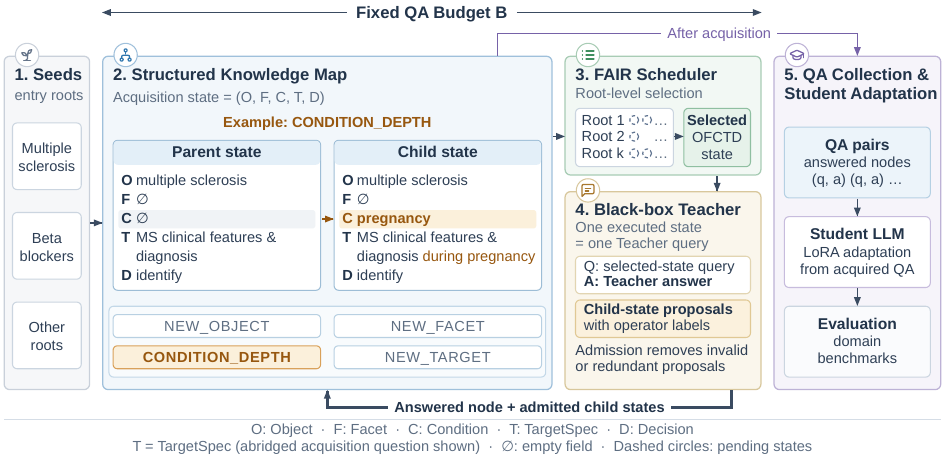}
    \caption{Overview of MapSteal. Frozen domain roots initialize a structured acquisition map. Root-level FAIR scheduling selects pending states; a deterministic realizer converts non-root states into questions. A black-box teacher returns an answer, explanation, and typed child-state proposals in one successful interaction. Admitted children re-enter the frontier, while collected QA supervision is used to adapt the student only after acquisition. Map edges indicate acquisition provenance rather than verified domain relations.}
    \label{fig:method_overview}
\vspace{-15pt}
\end{figure*}

We evaluate MapSteal across multiple student
scales and knowledge domains. The results show
that structured recursive acquisition provides
useful downstream supervision under limited QA
budgets. Controlled ablations on Medicine show
that the full structured formulation improves
the three-benchmark average by 1.17 percentage
points over a simplified Object--Facet variant,
while removing condition-based expansion reduces
performance by 1.05 points. Additional analyses
examine sensitivity to initial acquisition roots
and characterize the trade-off between recursive
depth, semantic novelty, and local coherence.

\noindent \textit{\textbf{Our contributions are threefold:}}
\begin{itemize}

\item \textbf{Explicit acquisition-state formulation.}
We formulate budget-constrained knowledge acquisition
as the construction and execution of an evolving
space of structured knowledge targets, distinguishing
query-space expansion from selection within
the available frontier.

\item \textbf{Structured recursive acquisition.}
We introduce MapSteal, which combines O/F/C/T/D
acquisition states, teacher-guided typed expansion,
proposal admission, and root-level fair scheduling
to support traceable recursive knowledge acquisition
without online student failure attribution.

\item \textbf{Empirical evaluation and analysis.}
We evaluate the framework across student scales
and domains, investigate its structured components
through controlled ablations, and characterize
the effects of acquisition roots, query budgets,
and recursive exploration behavior.

\end{itemize}


\section{Methodology}
\label{sec:method}
\noindent\textit{Overview.}
MapSteal acquires domain-oriented supervision from a black-box teacher under a finite query budget. Instead of diagnosing student errors online, it maintains an explicit frontier of structured knowledge-acquisition targets. Each successful teacher interaction answers the selected target and proposes possible successors. The successors are validated, represented as new states, and scheduled for later execution. The student is adapted only after acquisition is complete. Figure~\ref{fig:method_overview} summarizes the procedure.

\subsection{Problem Formulation}
\label{sec:problem}

Let $\mathcal{T}$ be a teacher language model accessible through a black-box query interface, and let $S$ be a student to be adapted to a target domain. The input is a frozen set of $K$ acquisition roots, $\mathcal{R}=\{r_1,\ldots,r_K\}$, each specifying an initial knowledge-seeking state and a question. Roots are domain entry points and need not be labeled downstream evaluation examples. Any fixed cold-start QA corpus $\mathcal{D}_{\mathrm{seed}}$ is distinct from $\mathcal{R}$ and is not part of the recursive acquisition loop.

The procedure produces an ordered sequence of successfully executed QA records,
\begin{equation}
\mathcal{D}_{\mathrm{acq}}=\big[(q_i,a_i,e_i)\big]_{i=1}^{N},
\qquad N\leq B_{\mathrm{acq}},
\label{eq:acquisition_data}
\end{equation}
where $a_i$ and $e_i$ are the teacher's answer and explanation for the executed question $q_i$. The acquisition budget $B_{\mathrm{acq}}$ counts successful QA executions, including root executions. API requests, invalid responses, and retries are accounted for separately. For experiments with cold-start examples, the nominal pre-QC QA budget is $B_{\mathrm{QA}}=|\mathcal{D}_{\mathrm{seed}}|+B_{\mathrm{acq}}$; the final training set may contain fewer rows after quality control (QC).
Our objective is to collect supervision that supports downstream student adaptation, not to establish complete coverage of a domain or replicate all teacher behavior. Crucially, the acquisition policy does not consult $S$, its predictions, or its uncertainty during execution. Downstream student performance measures the usefulness of acquired supervision after the acquisition process has ended.

\subsection{Structured Knowledge Acquisition Map}
\label{sec:acquisition_map}

Each acquisition target is represented by an explicit schema consisting of five components,
\begin{equation}
 s=(O,F,C,T,D),
 \label{eq:ofctd_state}
\end{equation}
where $O$ identifies a knowledge \emph{Object}, $F$ its \emph{Facet}, $C$ a contextual \emph{Condition}, $T$ a \emph{TargetSpec} describing the information sought, and $D$ the corresponding \emph{Decision} intent. Optional fields may be empty. These fields specify an acquisition request, not verified domain facts; in particular, $T$ is distinct from the final surface-form question.

MapSteal represents acquisition targets as
structured records rather than isolated
generated questions.
We define the acquisition map as
$\mathcal{M}_t=(V_t,E_t)$, where $V_t$
contains admitted acquisition states and
$E_t$ records directed parent--child links.
Each state maintains its source root and
parent provenance, allowing recursive
expansion to be traced explicitly.  A single ordered frontier $\mathcal{F}_t\subseteq V_t$ contains states awaiting execution; successfully executed states are recorded in the acquisition trajectory. Rejected proposals remain available in audit records with their rejection reasons but are not placed on the executable frontier. Neither the graph edges nor the schema assert ontological, causal, or otherwise verified relationships between domain entities.
For example, a target concerning multiple-sclerosis clinical features and diagnosis may produce a child concerning the same target during pregnancy. The child preserves the knowledge object while changing its condition and target specification. This transition is recorded as acquisition provenance. A syntactically valid combination of schema fields is not automatically a useful target: new states enter the frontier through teacher proposals and explicit admission checks, not exhaustive cross-products or object--object joins.

\subsection{Teacher-Guided Recursive Expansion}
\label{sec:recursive_expansion}

\paragraph{Question realization and teacher interaction.}
For a selected root state, MapSteal uses the question supplied in the frozen root manifest. For a non-root state, a deterministic realizer converts its TargetSpec and Decision intent into a knowledge-seeking question, including condition context when needed. Realization failure causes the state to be discarded before any teacher call. The prompt for a successfully realized state includes the current structured state and root-specific direction-registry context. In one successful teacher interaction, $\mathcal{T}$ returns an answer, an explanation, and typed child-state proposals; the teacher proposes \emph{states}, rather than directly supplying the final surface questions for those children. The direction registry records prior teacher-proposed directions for that root and informs the prompt, but not the scheduler.

\paragraph{Typed state transitions.}
Each proposed child is constructed using one of four operators. Table~\ref{tab:mapsteal_operators} gives their field-level effects. Every operator supplies a new TargetSpec and Decision intent; an inherited Condition is a field of the state, not a cross-root resource.
\textsc{ConditionDepth} changes the contextual
Condition while preserving the Object and Facet.
The new Condition may introduce, replace, or
refine constraints; the operator
does not require strict semantic specialization.

\begin{table}[thbp]
\vspace{-5pt}
\centering
\caption{Field updates for proposed states. ``Keep'' means inherit the parent's field; ``New'' means use the proposed value. Transitions assign new $T$ and $D$.}
\label{tab:mapsteal_operators}
\small
\setlength{\tabcolsep}{3.5pt}
\vspace{-5pt}
\begin{tabular}{lcccc}
\toprule
\textbf{Operator} & $O$ & $F$ & $C$ & $T,D$ \\
\midrule
\textsc{NewObject}       & New  & Empty & Empty & New \\
\textsc{NewFacet}        & Keep & New   & Keep  & New \\
\textsc{ConditionDepth}  & Keep & Keep  & New   & New \\
\textsc{NewTarget}       & Keep & Keep  & Keep  & New \\
\bottomrule
\end{tabular}
\vspace{-15pt}
\end{table}

\paragraph{Validation and admission.}
The interaction has three distinct safeguards. First, the returned teacher response must satisfy its schema; an invalid response may be retried once with the same request. Second, proposed children undergo operator-specific validity and admission checks, including required fields, valid Decision intents, parent/ancestor TargetSpec constraints, Decision--TargetSpec compatibility, and duplicate identities. Deduplication uses normalized exact matching, including across roots, rather than embedding similarity. Third, question realization is checked only when an admitted child is later selected for execution. Proposal-level rejection does not consume a successful QA budget slot. Accepted children are appended to the frontier and linked to their executed parent; rejected proposals and reasons remain in the trajectory audit. This separation between proposing, admitting, and executing targets makes recursive acquisition explicit and traceable.

\subsection{Budgeted Acquisition and Scheduling}
\label{sec:budgeted_scheduling}

Recursive proposals can accumulate faster than queries are executed. MapSteal therefore distinguishes the global ordered frontier $\mathcal{F}_t$ from the smaller set of states that can be executed under the remaining budget. Let $\mathcal{R}_t^{+}$ denote roots that have at least one pending state, and let $E_r$ be the number of \emph{successfully executed non-root} states associated with root $r$. The scheduler selects
\begin{equation}
 r^*=\underset{r\in\mathcal{R}_t^{+}}{\operatorname{argmin}}
 \big(E_r,\tau_r,\operatorname{id}(r)\big),
 \label{eq:fair_scheduler}
\end{equation}
using lexicographic order, where $\tau_r$ is the global-frontier position of the oldest pending state for $r$. It then executes that root's oldest pending state (FIFO). The implementation uses one global frontier, rather than physically separate queues. Because roots are initially enqueued before their descendants and root executions do not increment $E_r$, the root states are executed before recursive descendants; FAIR primarily balances subsequent recursive executions across roots.
Algorithm~\ref{alg:mapsteal} gives the core acquisition loop. Invalid realization or an invalid teacher response after its permitted retry removes the selected state without adding a successful QA record. If the frontier is exhausted, acquisition aborts rather than inventing replacement states. The actual runner additionally enforces cost and validity safeguards, detailed in the implementation appendix. Neither the scheduling rule nor the root-specific direction registry estimates downstream student utility or semantic coverage.

\begin{algorithm}[thbp]
\caption{MapSteal recursive acquisition}
\label{alg:mapsteal}
\small
\begin{algorithmic}[1]
\Require Frozen roots $\mathcal{R}$, teacher $\mathcal{T}$, budget $B_{\mathrm{acq}}$
\Ensure Ordered acquired QA records $\mathcal{D}_{\mathrm{acq}}$
\State $\mathcal{F}\gets\mathcal{R}$; $\mathcal{D}_{\mathrm{acq}}\gets[\,]$
\State Initialize global identities $\mathcal{V}$, counts $E_r\gets0$, registries $G_r$
\While{$|\mathcal{D}_{\mathrm{acq}}|<B_{\mathrm{acq}}$}
  \If{$\mathcal{F}=\emptyset$} \State \textbf{abort} \EndIf
  \State $r\gets\operatorname{FAIR}(\mathcal{F},E)$
  \State $s\gets\operatorname{PopOldest}(\mathcal{F},r)$
  \State $q\gets\operatorname{RootQuestion}(s)$ if root;
  \Statex \hspace{\algorithmicindent}\hspace{\algorithmicindent}otherwise $\operatorname{Realize}(s)$
  \If{$q=\bot$} \State \textbf{continue} \EndIf
  \State $y\gets\operatorname{QueryTeacher}(\mathcal{T},s,q,G_r)$
  \Statex \hspace{\algorithmicindent}\hspace{\algorithmicindent}\Comment{Response schema check; at most one retry}
  \If{$y=\bot$} \State \textbf{continue} \EndIf
  \State Append $(q,y.\mathrm{answer},y.\mathrm{explanation})$ to $\mathcal{D}_{\mathrm{acq}}$
  \State $\mathcal{C}\gets\operatorname{ExpandAndAdmit}(s,y,\mathcal{V})$
  \State Update $G_r$, $\mathcal{V}$, and append $\mathcal{C}$ to $\mathcal{F}$
  \If{$\operatorname{depth}(s)>0$} \State $E_r\gets E_r+1$ \EndIf
  \State Record executed state and proposal provenance
\EndWhile
\State \Return $\mathcal{D}_{\mathrm{acq}}$
\end{algorithmic}
\end{algorithm}

\subsection{Student Adaptation}
\label{sec:student_adaptation}

Each successfully acquired record contains a question, answer, and explanation. Offline QC removes malformed supervision and duplicate question texts. If cold-start supervision is supplied, it precedes acquired records in the QC input; consequently, a later acquired duplicate of an existing seed question is removed. Writing $\mathbin{\|}$ for ordered corpus concatenation, the final training corpus is
\begin{equation}
 \mathcal{D}_{\mathrm{train}}=
 \operatorname{QC}\!\left(\mathcal{D}_{\mathrm{seed}}\mathbin{\|}\mathcal{D}_{\mathrm{acq}}\right).
 \label{eq:train_corpus}
\end{equation}
The training target uses the teacher's answer followed by its explanation; ancillary proposal metadata, including the teacher's reusable-knowledge field, does not enter the training text. QC does not trigger replacement queries to restore the nominal budget. The student is adapted only after the full acquisition run:
\begin{equation}
 S'=\operatorname{FineTune}(S,\mathcal{D}_{\mathrm{train}}).
 \label{eq:student_update}
\end{equation}
Training configurations and downstream evaluation protocols are specified in the experimental setup. We evaluate the transferred utility of acquired supervision through student performance, while separately analyzing acquisition behavior (e.g., recursion depth and query similarity). An expanding frontier alone is not evidence of complete domain coverage or improved student performance.

\begin{table*}[thbp]
\centering
\caption{
Main results across selected medical and cybersecurity benchmarks.
Results are reported as accuracy (\%).
}
\label{tab:main_results}

\resizebox{\textwidth}{!}{
\begin{tabular}{llcccccc}
\toprule
& &
\multicolumn{3}{c}{\textbf{Medical}} &
\multicolumn{3}{c}{\textbf{Cybersecurity}} \\
\cmidrule(lr){3-5}
\cmidrule(lr){6-8}

\textbf{Student}
& \textbf{Method}
& \textbf{MedMCQA}
& \textbf{MedQA}
& \textbf{MedXpert}
& \textbf{CyberMetric}
& \textbf{SecEval}
& \textbf{CTI-Bench} \\
\midrule

\multirow{4}{*}{Qwen3.5-2B}
& Pristine
& 45.80 & 36.06 & 10.78
& 79.47 & 71.93 & 60.64 \\

& EvoKD
& $40.53\pm8.09$
& $41.19\pm8.47$
& $12.39\pm1.57$
& $71.00\pm6.19$
& $69.49\pm6.25$
& $52.73\pm9.42$ \\

& QEDKS
& $43.27\pm0.12$
& $44.59\pm0.81$
& $\mathbf{14.88\pm0.83}$
& $79.49\pm0.23$
& $77.73\pm1.46$
& $\mathbf{65.48\pm0.16}$ \\

& \textbf{MapSteal (Ours)}
& $\mathbf{47.87{\pm}1.03}$
& $\mathbf{45.59{\pm}0.16}$
& $14.65{\pm}0.19$
& $\mathbf{79.70{\pm}0.14}$
& $\mathbf{78.68{\pm}0.56}$
& $62.41{\pm}1.07$ \\

\midrule

\multirow{4}{*}{Qwen3.5-4B}
& Pristine
& 56.60 & 62.14 & 16.73
& 83.63 & 80.60 & 65.71 \\

& EvoKD
& $58.47\pm1.10$
& $63.45\pm0.66$
& $16.22\pm0.70$
& $82.97\pm0.13$
& $80.28\pm0.75$
& $63.79\pm0.93$ \\

& QEDKS
& $61.00\pm0.53$
& $65.54\pm0.20$
& $16.94\pm0.45$
& $84.21\pm0.24$
& $81.46\pm0.96$
& $\mathbf{67.30\pm0.54}$ \\

& \textbf{MapSteal (Ours)}
& $\mathbf{62.27\pm0.42}$
& $\mathbf{66.69\pm0.44}$
& $\mathbf{17.63\pm0.46}$
& $\mathbf{84.61\pm0.16}$
& $\mathbf{83.19\pm0.33}$
& $66.59\pm0.71$ \\

\midrule

\multirow{4}{*}{Qwen3.5-9B}
& Pristine
& 62.60 & 67.40 & 19.55
& 83.75 & 83.04 & 64.99 \\

& EvoKD
& $59.00\pm5.72$
& $67.69\pm5.64$
& $19.85\pm0.42$
& $84.10\pm1.00$
& $81.00\pm2.22$
& $66.95\pm1.73$ \\

& QEDKS
& $65.13\pm0.90$
& $\mathbf{71.20\pm0.71}$
& $19.88\pm0.50$
& $\mathbf{86.75\pm0.14}$
& $84.18\pm0.40$
& $\mathbf{70.52\pm0.19}$ \\

& \textbf{MapSteal (Ours)}
& $\mathbf{65.60{\pm}0.20}$
& $71.14{\pm}0.43$
& $\mathbf{20.18{\pm}0.08}$
& $86.66{\pm}0.12$
& $\mathbf{85.32{\pm}0.43}$
& $69.04{\pm}0.77$ \\

\bottomrule
\end{tabular}
}

\vspace{2pt}
\footnotesize{
$\pm$ denotes the sample standard deviation over three seeds.
}
\vspace{-4pt}

\end{table*}

\section{Experiments}
\label{sec:experiments}

\subsection{Setup}
\label{sec:experimental_setup}
\paragraph{Models and Training.}
We use GPT-4.1 as the black-box Teacher and Qwen3.5-2B,
4B, and 9B as Students~\citep{qwen35blog}.
Students are adapted using LoRA~\citep{hu2021lora}
after knowledge acquisition. Unless otherwise noted,
we report mean $\pm$ standard deviation over three
seeds. Further implementation
details are provided in Appendix~\ref{app:implementation}.

\paragraph{Benchmarks and Evaluation.}
We evaluate knowledge transfer in two domains:
Medicine and Cybersecurity.
For Medicine, we use MedMCQA~\citep{medmcqa},
MedQA~\citep{medqa}, and
MedXpert~\citep{medxpertqa}.
For MedMCQA, we evaluate on a fixed subset of
500 questions from the official validation
(dev) split, as the test labels are not
publicly available.
For Cybersecurity, we use
CyberMetric~\citep{CyberMetric},
SecEval~\citep{li2023seceval}, and
CTI-Bench~\citep{alam2024ctibenchbenchmarkevaluatingllms}.
We report classification or multiple-choice
accuracy (\%) on the corresponding evaluation
splits. For domain-level comparisons, we compute
the unweighted mean across the three benchmarks
within each domain. This aggregation gives equal
weight to each benchmark rather than to each
individual evaluation question.

\paragraph{Baselines.}
We compare MapSteal with the following methods.
\textit{Pristine} denotes the original Student
checkpoint without task-specific adaptation.
We additionally compare against
\textit{EvoKD}~\citep{EvoKD} and
\textit{QEDKS}~\citep{Query-efficientdomainknowledgestealing}, which serve
as knowledge acquisition and distillation baselines.
\textit{MapSteal} is our structured,
Teacher-guided recursive acquisition method.
All adapted methods are evaluated with the same
Student model family and corresponding benchmark
protocols.

\paragraph{Acquisition Budget.}
We use a nominal budget of $B=500$ QA supervision instances.
Both predefined root questions and recursively
generated questions require Teacher responses
and count toward this budget. Further protocol
details are provided in
Appendix~\ref{app:implementation}. 
Teacher token usage and estimated acquisition costs
are reported in Appendix~\ref{app:teacher_cost}.

\subsection{Main Results}
\label{sec:main_results}

Table~\ref{tab:main_results} compares MapSteal with
Pristine, EvoKD, and QEDKS under a nominal budget
of 500 QA supervision instances across three Student
scales and two domains.
MapSteal achieves the highest mean accuracy on
12 of the 18 evaluated settings, with particularly
consistent results in Medicine. For the 4B Student,
MapSteal outperforms both acquisition baselines on
all three medical benchmarks, reaching 62.27\%
on MedMCQA, 66.69\% on MedQA, and 17.63\% on
MedXpertQA. MapSteal also achieves competitive results
at 2B and 9B, although QEDKS obtains the
highest MedQA accuracy at 9B.
In Cybersecurity, MapSteal achieves the highest
CyberMetric and SecEval accuracy at 4B, while
QEDKS performs better on CTI-Bench. At 9B, QEDKS
leads on CyberMetric and CTI-Bench, whereas
MapSteal achieves the highest SecEval accuracy.
These results indicate that the relative benefits
of acquisition methods vary across benchmarks
and Student scales.

EvoKD exhibits substantial performance degradation
and high run-to-run variance in several settings,
particularly with the 2B Student. All three runs
are retained in the reported results.
Per-run results and further analysis are provided
in Appendix~\ref{app:evokd_variance}.
Overall, MapSteal achieves competitive downstream
performance across the evaluated settings,
with the highest mean accuracy in 12 of 18
Student--benchmark combinations.
These results demonstrate that Student-independent
recursive acquisition can provide useful
supervision for downstream adaptation.
We further examine the effect of acquisition data size
under different QA budgets, with results
provided in Appendix~\ref{app:budget_scaling}.

\subsection{Ablation Study}

We investigate three aspects of MapSteal:
the contribution of structured acquisition,
sensitivity to the number of initial acquisition roots,
and the effect of increasing the acquisition budget.
Unless otherwise stated, experiments use
Qwen3.5-4B on Medicine with the same
Student adaptation configuration.

\subsubsection{Effect of Structured Acquisition}

\begin{table*}[htbp]
\vspace{-10pt}
\centering
\caption{Ablation study of MapSteal on medical benchmarks.}
\label{tab:mapsteal_ablation}

\small
\setlength{\tabcolsep}{3.5pt}
\begin{tabular*}{\textwidth}{@{\extracolsep{\fill}}lccccc@{}}
\toprule
\textbf{Variant}
& \textbf{MedMCQA}
& \textbf{MedQA}
& \textbf{MedXpert}
& \textbf{Avg}
& $\boldsymbol{\Delta}$ \\
\midrule

\textbf{Full MapSteal}
& $\mathbf{62.27\pm0.42}$
& $\mathbf{66.69\pm0.44}$
& $\mathbf{17.63\pm0.46}$
& $\mathbf{48.86\pm0.17}$
& -- \\

w/o CONDITION\_DEPTH
& $60.93\pm0.31$
& $66.20\pm0.28$
& $16.30\pm0.14$
& $47.81\pm0.04$
& $-1.05$ \\

Simplified OF
& $60.87\pm0.31$
& $65.28\pm0.36$
& $16.94\pm0.32$
& $47.69\pm0.10$
& $-1.17$ \\

\bottomrule
\end{tabular*}
\vspace{-10pt}
\end{table*}

\begin{figure}[thbp]
\vspace{-11pt}
\centering
\makebox[\columnwidth][c]{%
    \includegraphics[width=1.08\columnwidth]{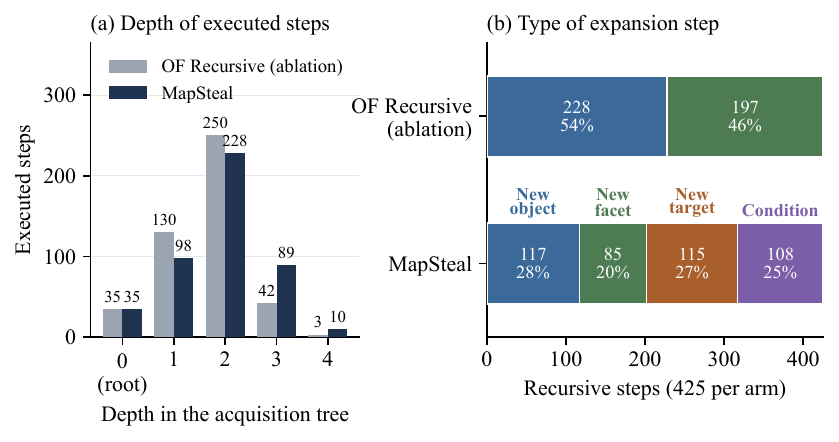}%
}
\caption{
Acquisition depth and expansion operator
distributions of Simplified OF and Full MapSteal.
Both methods execute 460 acquisition questions
from the same 35 roots.
Operator counts exclude root executions.
}
\label{fig:acquisition_coverage}
\vspace{-15pt}
\end{figure}

We compare Full MapSteal with two variants
under a fixed budget of 500 QA pairs,
comprising 40 shared cold-start examples and
460 Teacher acquisition executions.
\textit{Simplified OF} replaces the complete OFCTD
formulation with Object--Facet acquisition.
The \textit{w/o CONDITION\_DEPTH} variant disables
condition-based recursive expansion while retaining
the remaining structured formulation.
In Table~\ref{tab:mapsteal_ablation},
both ablations reduce downstream performance.
Replacing OFCTD with the simplified OF formulation
reduces average by 1.17 percentage points.
Disabling CONDITION\_DEPTH yields a similar
decrease of 1.05.
These findings support the complete structured
acquisition formulation and the contribution of
condition-based recursive expansion.
However, Simplified OF changes multiple acquisition
components, including state representation, question
realization, and proposal constraints.
The reported confidence intervals quantify
evaluation-item uncertainty conditional on the
observed acquisition trajectories.
Figure~\ref{fig:acquisition_coverage} further
compares the acquisition trajectories.
Full MapSteal produces more depth-3 and depth-4
executions than Simplified OF and distributes
recursive expansion across four operator types.

\subsubsection{Sensitivity to Initial Root Count}
We investigate how the number of initial acquisition
roots affects downstream performance under a fixed
QA budget. All variants share 40 cold-start QA pairs
and perform 460 additional Teacher acquisitions,
yielding 500 QA pairs before quality control.
We vary the number of acquisition roots over
$\{1,10,20,35\}$ using nested subsets of a fixed
35-root pool. The 35-root configuration corresponds
to Full MapSteal.

\begin{table}[t]
\vspace{-4pt}
\centering
\caption{
Effect of initial acquisition root count.
}
\label{tab:root_count}
\vspace{-4pt}
\small
\setlength{\tabcolsep}{3pt}
\begin{tabular}{cccc}
\toprule
\textbf{Roots}
& \textbf{MedMCQA}
& \textbf{MedQA}
& \textbf{MedXpert}
\\
\midrule

1
& $60.87\pm1.14$
& $66.25\pm0.67$
& $16.97\pm0.29$ \\

10
& $61.47\pm0.31$
& $65.49\pm0.39$
& $17.36\pm0.37$\\

20
& $61.27\pm0.50$
& $66.35\pm0.37$
& $16.87\pm0.36$\\

\textbf{35}
& $\mathbf{62.27\pm0.42}$
& $\mathbf{66.69\pm0.44}$
& $\mathbf{17.63\pm0.46}$\\

\bottomrule
\vspace{-20pt}
\end{tabular}
\end{table}
As shown in Table~\ref{tab:root_count}, MapSteal
achieves similar Medical3 scores with 1, 10, and 20
roots (48.03, 48.11, and 48.16, respectively).
The full 35-root configuration achieves 48.86,
outperforming these variants by 0.70--0.84
percentage points. The results show little
performance variation between 1 and 20 roots
in the evaluated configurations.
We also observe a clear difference in acquisition
depth. With the same query budget, mean trajectory
depth decreases from 5.27 with one root to 2.86,
2.37, and 1.87 with 10, 20, and 35 roots,
respectively. This suggests that fewer initial
roots concentrate acquisition into deeper
recursive trajectories, while additional roots
allow exploration to begin from more starting points.
Under the evaluated nested root subsets,
MapSteal retains much of its downstream
performance even with fewer initial roots.
Root subsets are nested subsets of a fixed 35-root pool, each with one Teacher acquisition trajectory.
Detailed acquisition depth and operator distributions are provided in Appendix~\ref{root_count_analysis}.

\subsection{Analysis of Acquisition Behavior}
We further analyze how
structured acquisition changes recursive exploration.
We analyze executed acquisition trajectories using depth, semantic novelty, and parent–child similarity.

\begin{table}[thbp]
\vspace{-9pt}

\centering
\caption{
Descriptive acquisition statistics for Simplified OF
and Full MapSteal. Semantic metrics are computed
from executed acquisition questions.
}
\label{tab:acquisition_statistics}
\vspace{-5pt}

\small
\setlength{\tabcolsep}{3pt}

\begin{tabular*}{\columnwidth}{@{\extracolsep{\fill}}lcc@{}}
\toprule
\textbf{Metric}
& \textbf{OF}
& \textbf{OFCTD} \\
\midrule

Mean trajectory depth
& 1.67 & 1.87 \\

Distinct Object strings
& 231 & 147 \\

Sequential novelty
& 0.262 & 0.178 \\

Parent--child cosine
& 0.621 & 0.788 \\

Near-duplicate rate (\%)
& 8.5 & 30.8 \\

\bottomrule
\end{tabular*}
\vspace{-18pt}
\end{table}
\subsubsection{Recursive Exploration Patterns}

We compare the acquisition behavior of Simplified OF
and Full MapSteal using trajectory depth, semantic
novelty, and parent--child similarity.
As shown in Table~\ref{tab:acquisition_statistics},
Simplified OF explores more distinct Object strings
and exhibits greater sequential novelty.
In contrast, Full MapSteal maintains higher
parent--child similarity and greater
mean recursive depth.
These results suggest different exploration patterns:
OF favors broader changes in acquisition objects,
whereas OFCTD produces trajectories with stronger
local semantic similarity.

\begin{figure*}[t]
\vspace{-15pt}

\centering
\includegraphics[width=0.98\textwidth]
{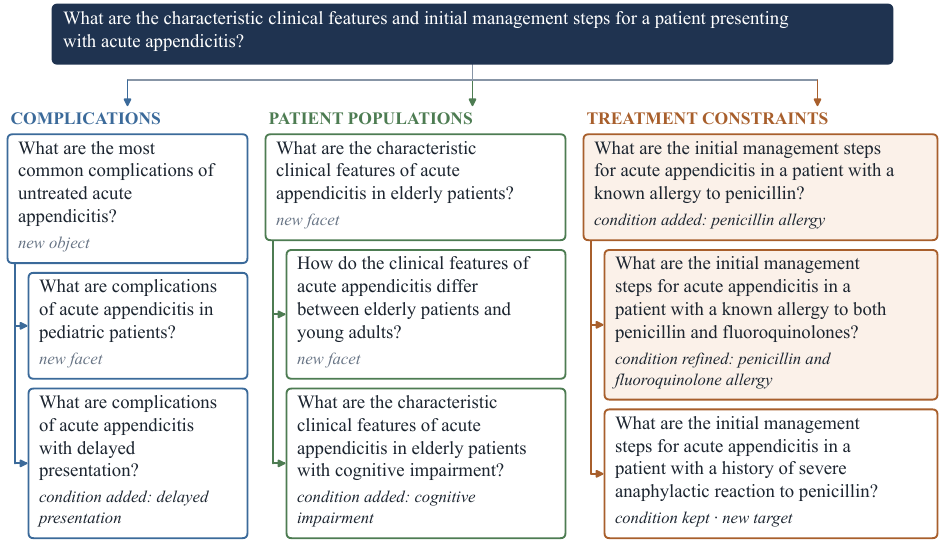}
\caption{
Qualitative example of structured recursive
knowledge acquisition.
Starting from an acute appendicitis question,
MapSteal expands into related directions involving
complications, patient populations, and treatment
constraints, followed by context-specific questions.
The figure displays 10 selected nodes from a real
14-node executed acquisition tree.
Questions are shown verbatim, and arrows represent
recorded parent--child links.
The complete tree is provided in
Appendix~\ref{app:appendicitis_complete}.
}
\label{fig:qualitative_acquisition}
\vspace{-16pt}
\end{figure*}

\noindent\textbf{Qualitative Example.}
Figure~\ref{fig:qualitative_acquisition} illustrates
how MapSteal recursively expands a general medical question
into multiple context-specific knowledge targets.
Starting from acute appendicitis, the acquisition process
explores complications, different patient populations,
and treatment constraints.
Follow-up questions introduce additional
contexts, including delayed presentation,
cognitive impairment in elderly patients,
and multiple drug allergies.
All displayed questions were generated and executed
without additional human-authored queries beyond
the initial roots.
This example illustrates coherent branching and
contextual specialization, rather than independently
validated medical knowledge.
Additional acquisition visualizations, including
the complete executed tree underlying
Figure~\ref{fig:qualitative_acquisition},
are provided in
Appendix~\ref{app:acquisition_visualizations}.
%







\section{Related Work}
\noindent \textbf{Black-Box Model Extraction and Distillation.}
Model extraction trains surrogate models through
black-box query access. Early studies demonstrate
functionality stealing for image classifiers
and NLP models using selected or synthetic
queries~\cite{KnockoffNets,thievessesamestreetmodel}.
Recent work extends this setting to black-box
LLMs. Model Leeching acquires task-specific
supervision~\cite{modelleechingextractionattack},
while LoRD~\cite{LoRD}, GAD, and SODA develop strategies
for improving Student adaptation from Teacher
responses
~\cite{GAD,sodasemionpolicyblackbox}
MapSteal focuses on the preceding acquisition
stage: how to autonomously construct and expand
knowledge targets before Student training.

\noindent \textbf{Query Selection and Generation.}
Query-efficient model extraction commonly uses
active learning to select informative samples,
as in ActiveThief and subsequent text-classification
methods~\cite{ActiveThief,Query-efficientmodelextractionfortextclassification}.
Data-free approaches instead synthesize queries
to expose differences between the victim and
surrogate~\cite{maze,datafreemodelextraction}.
More recent methods iteratively discover domain
entities, refine questions using Student feedback,
or maintain graph-based exploration
states~\cite{Query-efficientdomainknowledgestealing,
queryefficientagenticgraphextraction}.
In contrast, MapSteal recursively expands
structured acquisition targets through
Teacher-proposed directions without relying
on Student uncertainty or failure signals.

\noindent \textbf{Adaptive Knowledge Acquisition.}
Iterative distillation methods adapt acquired
supervision as Student capabilities evolve.
EvoKD generates new examples from Student
predictions~\cite{EvoKD}, while INTERACT supports
multi-turn acquisition of unseen
concepts~\cite{interact}.
Related approaches use uncertainty, model
errors, or capability-driven task generation
to guide further learning
~\cite{activesyntheticdatageneration,r-zero,diagevo}.
State-aware information-gathering methods,
including BED-LLM and DualGraph, also organize
exploration through explicit belief or knowledge
states~\cite{bedllmintelligentinformationgathering,
talegraphsseparatingknowledge, graphsub}.
MapSteal instead maintains structured acquisition
states and expands Teacher-proposed targets
from predefined roots, separating autonomous
knowledge acquisition from subsequent Student
adaptation.

\section{Conclusion}
We introduced MapSteal, a student-independent
framework that recursively expands structured
knowledge targets under a fixed QA budget.
By explicitly representing acquisition states
and their relationships, MapSteal organizes
teacher-guided exploration without requiring
Student feedback during acquisition.
Its structured operators enable acquisition
to branch across knowledge objects, facets,
and contextual conditions while preserving
the provenance of generated questions.
Across three Student scales and two domains,
MapSteal achieves the highest mean accuracy
in 12 of 18 settings, including three
medical benchmarks at the 4B scale and the
highest SecEval accuracy at every Student scale.
Ablations show that replacing the full
formulation with a simplified Object--Facet variant
lowers the three-benchmark average by 1.17 points,
and removing condition-based expansion lowers it
by 1.05 points.
Varying the number of initial roots changes
recursive depth substantially, from a mean of 5.27
with one root to 1.87 with 35, while downstream
accuracy remains close across 1 to 20 roots.
Increasing the QA budget from 40 to 500 raises
the Medical3 average from 47.27 to 48.86.
Qualitative analyses further illustrate how
recursive acquisition generates context-specific
knowledge targets from a limited set of roots.
Cross-domain gains vary by domain, motivating
future work on assessing and prioritizing
acquisition targets to improve the effectiveness
of recursive knowledge transfer.

\clearpage

\section*{Limitations}
\label{sec:limitations}

\paragraph{Acquisition variability.}
GPT-4.1 responses may vary across API calls even
under identical prompts and settings, so identical
roots and configurations do not guarantee identical
acquisition trajectories.
We use one Teacher acquisition trajectory per
configuration with multiple Student training seeds,
so the reported standard deviations do not capture
variability across independent acquisition runs.

\paragraph{Acquisition quality and scope.}
Our acquisition analysis measures structural and
semantic properties of generated questions, not
independently verified knowledge quality or domain
coverage.
The main evaluation covers medical and cybersecurity
benchmarks with Qwen3.5, and cross-domain
gains vary by domain and evaluation panel.

\section*{Ethical Considerations}
\label{sec:ethical_considerations}

This work studies structured knowledge acquisition
from black-box language models for model distillation
and small-model adaptation.
Such techniques can support reproducible research,
but they also present dual-use risks: recursive
acquisition could be used to imitate capabilities of
proprietary models without authorization.
The use and redistribution of Teacher outputs may be
subject to API terms, licensing conditions, and
intellectual property restrictions, and researchers
applying these methods should respect applicable
access permissions, service terms, and data-use
requirements.
Our experiments include medical and cybersecurity
domains, where incorrect acquired supervision could
produce unreliable downstream predictions; the
resulting Student models should not be treated as
clinically validated or as suitable for
security-critical deployment without further
evaluation.
We release implementation materials to support
reproducibility and evaluation, and we encourage
responsible access to and use of black-box model
interfaces.

\bibliography{custom}

\clearpage
\appendix

\appendix

\section{Implementation and Evaluation Details}
\label{app:implementation}
\label{sec:appendix}

\subsection{Models and Optimization}
MapSteal uses GPT-4.1 (\texttt{gpt-4.1-2025-04-14})
as the Teacher with temperature 0. Students are
adapted using LoRA~\citep{hu2021lora}, with
hyperparameters summarized in Table~\ref{tab:lora_config}.

\begin{table}[h]
\centering
\small
\begin{tabular}{lc}
\toprule
\textbf{Hyperparameter} & \textbf{Value} \\
\midrule
LoRA rank & 16 \\
LoRA alpha & 32 \\
LoRA dropout & 0.05 \\
Target modules & gate, up, down projections \\
Learning rate & $3\times10^{-4}$ \\
Training epochs & 2 \\
Training seeds & 42, 43, 44 \\
\bottomrule
\end{tabular}
\caption{LoRA hyperparameters used for MapSteal Student adaptation.}
\label{tab:lora_config}
\vspace{-15pt}
\end{table}

\subsection{Implementation and Computational Resources}

All Student-side experiments used single-GPU jobs
on NVIDIA A100 40GB GPUs.
MapSteal performs acquisition through the GPT-4.1
API without Student-side GPU computation.

\begin{table}[t]
\centering
\caption{
Recorded GPU time for the final experiments.
Acquisition and online-training durations include
Teacher API waiting time. LoRA training required 15--19 minutes per seed
for 2B, 20--26 minutes for 4B, and 20--25 minutes
for 9B Students.
Final evaluation covered all 54 adapters,
requiring 7.3, 7.7, and 7.4 GPU-hours for
MapSteal, QEDKS-Adapted, and EvoKD-Adapted,
respectively.
}
\label{tab:gpu_resources}
\small
\setlength{\tabcolsep}{4pt}
\begin{tabular}{lrr}
\toprule
\textbf{Stage} & \textbf{Runs} & \textbf{Time (h)} \\
\midrule
MapSteal LoRA training & 18 & 6.3 \\
QEDKS LoRA training & 18 & 6.6 \\
QEDKS acquisition & 18 & $\sim$36.0 \\
EvoKD online acquisition/training & 18 & $\sim$54.0 \\
\midrule
Final evaluation (all methods) & 54 & 22.4 \\
\bottomrule
\end{tabular}
\end{table}

Training and evaluation times are measured from
run logs, while acquisition and online-training
durations are recorded job wall-clock times
that include Teacher API latency.
The evaluation total excludes pristine-model
evaluation and additional model-loading overhead.
Teacher token usage and estimated API costs
are reported in Appendix~\ref{app:teacher_cost}.

\subsection{Acquisition Budget and Quality Control}
We use a nominal budget of $B=500$ Teacher QA
examples, including predefined root questions
and recursively generated questions. Root
questions are prepared before acquisition but
still require Teacher responses. Each successful
QA execution counts toward the budget regardless
of how its question was constructed.
For Medicine, the supervision pool contains
40 previously acquired seed QA examples and
460 new Teacher QA executions, comprising
35 root questions and 425 recursive questions.
The 40 seed examples count toward the total
supervision budget, although they do not incur
new Teacher calls in the current acquisition run.
For Cybersecurity, all 500 QA examples are
obtained during acquisition.
After quality control, five Medicine questions
duplicate existing seed examples, leaving
495 training records.


\subsection{Randomness and Baseline Runs}
MapSteal reports three Student training seeds on a fixed
Teacher acquisition trajectory. EvoKD reports three independent
acquisition and online adaptation runs. QEDKS reports three
independent acquisition runs, each paired with its matching
Student training seed. These repetitions capture different
sources of variability and should not be treated as equivalent.

\section{Additional Acquisition Analyses}
\label{app:additional_acquisition}

\subsection{Root Count Analysis}
\label{root_count_analysis}

We analyze how the number of initial acquisition
roots affects the structure of recursive exploration.
We consider the same four configurations used in the
root count ablation: R1, R10, R20, and R35.
The root subsets are nested, and the remaining
acquisition budget is allocated to recursive expansion.

\begin{figure*}[t]
\vspace{-15pt}
\centering
\IfFileExists{Figure/Root_count_depth_distribution.pdf}{
    \includegraphics[width=0.8\textwidth]
    {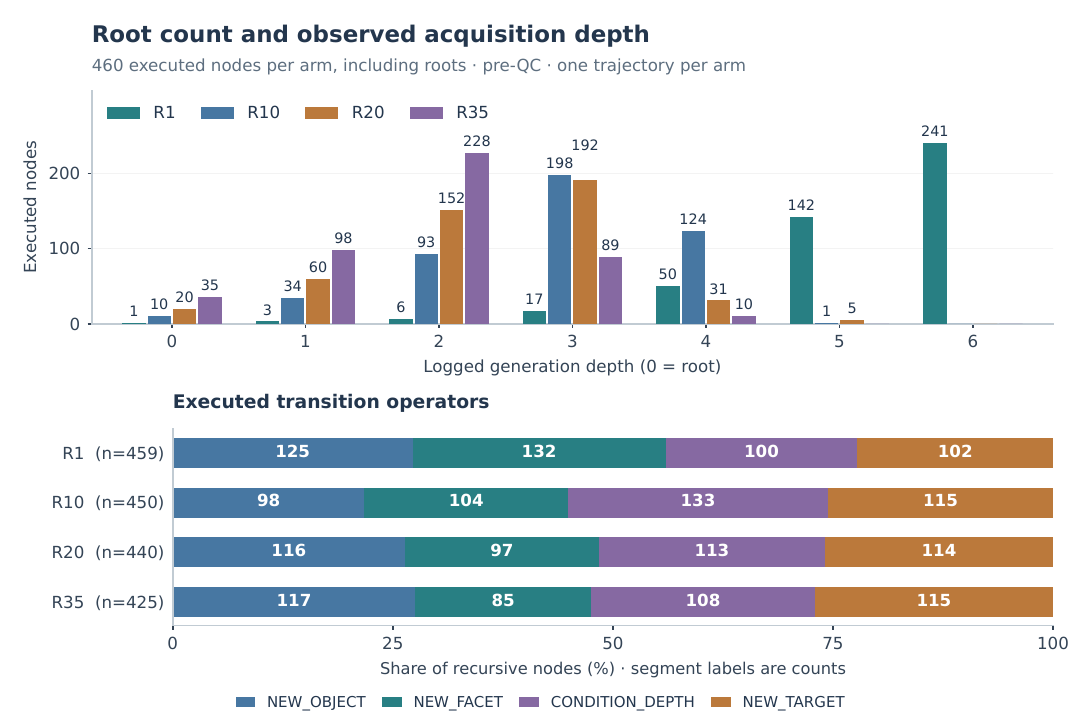}
}{
    \fbox{
    \parbox[c][2.0in][c]{0.9\textwidth}{
    \centering
    Root Count Depth and Operator Distribution

    \texttt{\detokenize{Figure/Root_count_depth_distribution.pdf}}
    }}
}
\caption{
Acquisition depth and expansion operator distributions
under different initial root counts.
Each configuration contains 460 executed acquisition
nodes. Depth statistics include the acquisition roots
but exclude the shared cold-start QA pairs.
}
\label{fig:root_count_depth}
\vspace{-15pt}
\end{figure*}

\begin{table}[t]
\centering
\caption{
Acquisition statistics under different initial
root counts. All configurations use 460 successful
Teacher acquisition queries.
}
\label{tab:root_count_statistics}

\small
\setlength{\tabcolsep}{4pt}
\begin{tabular}{lcccc}
\toprule
\textbf{Metric}
& \textbf{R1}
& \textbf{R10}
& \textbf{R20}
& \textbf{R35} \\
\midrule
Recursive nodes
& 459 & 450 & 440 & 425 \\
Mean depth
& 5.27 & 2.86 & 2.37 & 1.87 \\
Max depth
& 6 & 5 & 5 & 4 \\
CONDITION\_DEPTH
& 100 & 133 & 113 & 108 \\
Non-empty Condition (\%)
& 37.8 & 42.8 & 33.0 & 29.1 \\
Training QA after QC
& 500 & 498 & 496 & 495 \\
Teacher cost (\$)
& 5.98 & 5.29 & 5.02 & 4.84 \\
\bottomrule
\end{tabular}
\vspace{-10pt}
\end{table}

\paragraph{Recursive exploration depth.}
Figure~\ref{fig:root_count_depth} and
Table~\ref{tab:root_count_statistics} show that
fewer initial roots produce deeper acquisition
trajectories under the fixed query budget.
Mean depth decreases from 5.27 with one root
to 2.86, 2.37, and 1.87 with 10, 20, and 35
roots, respectively.
Similarly, the maximum observed depth decreases
from 6 to 4.
This pattern reflects a difference in exploration
structure. With fewer starting points, more
recursive queries are concentrated within a
smaller number of acquisition trees.
With additional roots, acquisition is distributed
across more independently initialized trees.

\paragraph{Expansion operators and conditions.}
The frequency of CONDITION\_DEPTH does not follow
a monotonic trend with root count.
The operator is executed 100, 133, 113, and 108
times for R1, R10, R20, and R35, respectively.
The proportion of executed states with non-empty
conditions also varies, reaching its highest value
of 42.8\% in R10.

These observations indicate that changing the
initial root set affects both the depth and the
types of recursive expansions.
However, operator frequency and condition usage
are descriptive properties of acquisition
trajectories, not direct measures of knowledge
coverage or downstream utility.

\paragraph{Acquisition cost and performance.}
Although all configurations execute the same number
of Teacher queries, their recorded costs differ.
R1 costs \$5.98, compared with \$4.84 for R35,
an increase of approximately 24\%.
The deeper R1 trajectory also requires more
input tokens, indicating that equal query counts
do not necessarily imply equal acquisition costs.
Despite the substantial differences in acquisition
structure, the downstream Medical3 scores of
R1, R10, and R20 remain close, ranging from
48.03 to 48.16, compared with 48.86 for R35
(Table~\ref{tab:root_count}).
Thus, deeper recursive exploration can maintain
much of the observed downstream performance,
but does not fully compensate for the difference
between these root configurations.

\subsection{Additional QA Budget Analysis}
\label{app:budget_scaling}
\begin{figure}[thbp]
  \centering
  \includegraphics[width=\columnwidth]{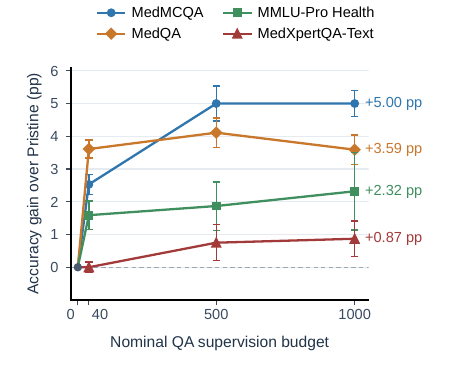}
  \caption{
  Extended QA budget scaling of MapSteal on Qwen3.5-4B
  for Medicine, under the LoRA recipe
  (learning rate $2\times10^{-4}$, three epochs,
  all seven LoRA target modules).
  The $y$-axis shows accuracy gain over the pristine Student
  in percentage points.
  Results show mean $\pm$ sample standard deviation
  over three training seeds using one fixed nested
  acquisition trajectory.
  The budget counts successfully acquired QA supervision,
  including the 40-example seed set, rather than API requests.
  B0 is the pristine Student without an adapter, and the
  dashed horizontal line marks zero gain.
  }
  \label{fig:extended_budget_scaling}
\end{figure}

We further examine the effect of QA supervision
budget on MapSteal using Qwen3.5-4B.
We report Medical3 accuracy averaged over
MedMCQA, MedQA, and MedXpertQA-Text
at five supervision budgets.
All checkpoints use nested prefixes of the same
acquisition trajectory and are independently
adapted from the pristine Student.
Results are reported as mean $\pm$ sample standard
deviation over three training seeds.
Medical3 accuracy increases from 47.27\% at B40
to 48.86\% at B500, an improvement of
1.59 percentage points.
Although intermediate results are not strictly
monotonic, the overall trend suggests that
additional acquisition supervision can improve
downstream performance.
\paragraph{Extended Budget Analysis.}
We additionally examine QA budget scaling under
a separate training configuration.
Figure~\ref{fig:extended_budget_scaling}
reports the corresponding results across
a wider range of supervision budgets.
The two experiments use different training
settings and are analyzed separately.

\section{Qualitative Acquisition Examples}
\label{app:acquisition_visualizations}
We provide qualitative visualizations of the
acquisition trajectories generated by Simplified OF
and Full MapSteal.
Both methods use the same 35 initial roots and
execute 460 acquisition queries.
We examine the global recursive structures,
a complete acquisition tree illustrating
context-specific question generation,
and an example of structured condition transitions.
All visualized nodes and edges are taken from
executed acquisition trajectories.

\subsection{Global Acquisition Structure}

\begin{figure*}[t]
\centering
\includegraphics[width=0.98\textwidth]
{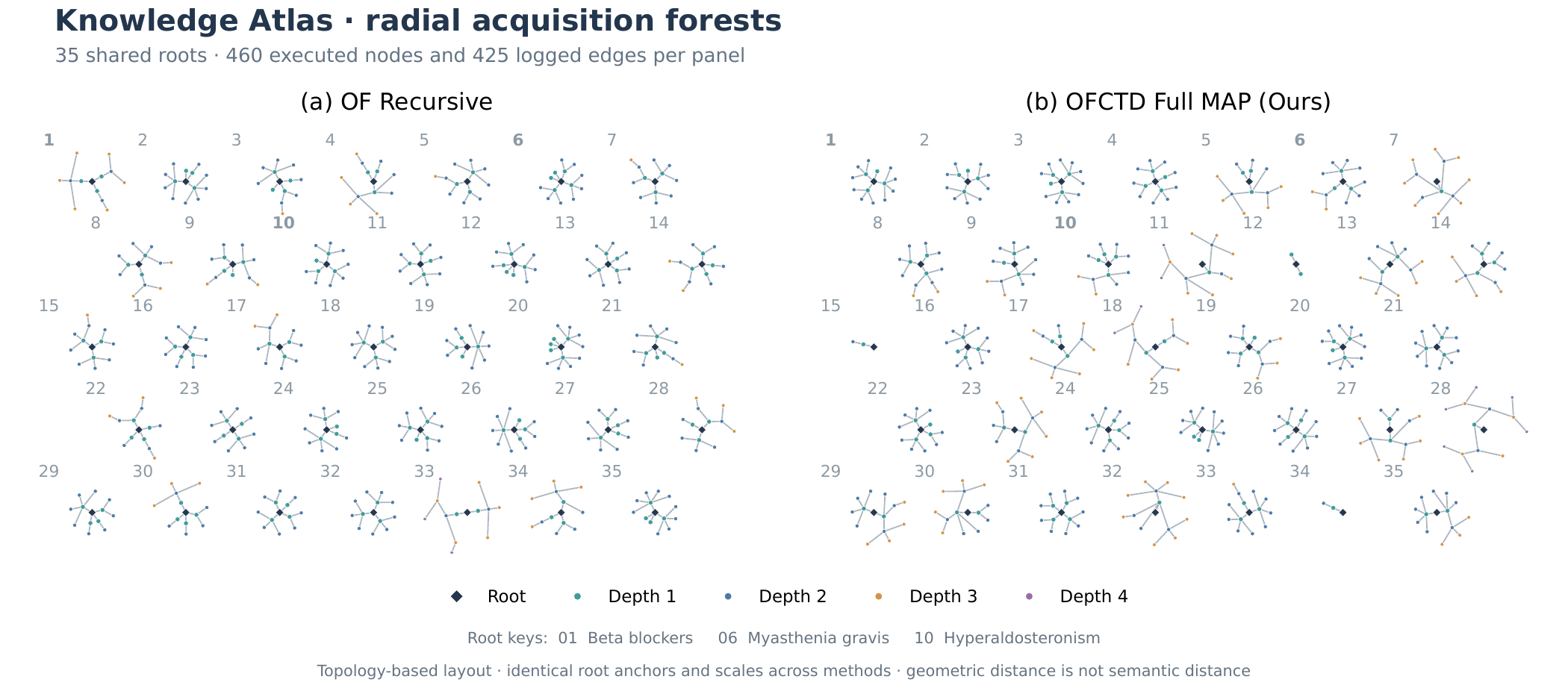}
\caption{
Complete acquisition forests of Simplified OF
(left) and Full MapSteal (right).
Both methods execute 460 acquisition questions
from the same 35 roots.
Nodes are colored by depth, and edges indicate
recorded parent--child relationships.
}
\label{fig:acquisition_forest}
\end{figure*}

Figure~\ref{fig:acquisition_forest} visualizes
the complete acquisition forests of Simplified OF
and Full MapSteal under the same roots and QA budget.
Each forest contains 35 roots, 460 executed nodes,
and 425 recorded parent--child edges.
Despite identical acquisition budgets, the methods
produce different recursive structures.
Node positions reflect graph topology and depth,
rather than semantic distance.

\subsection{Complete Appendicitis Acquisition Tree}
\label{app:appendicitis_complete}

\begin{figure*}[t]
\vspace{-10pt}
\centering
\includegraphics[width=0.8\textwidth]
{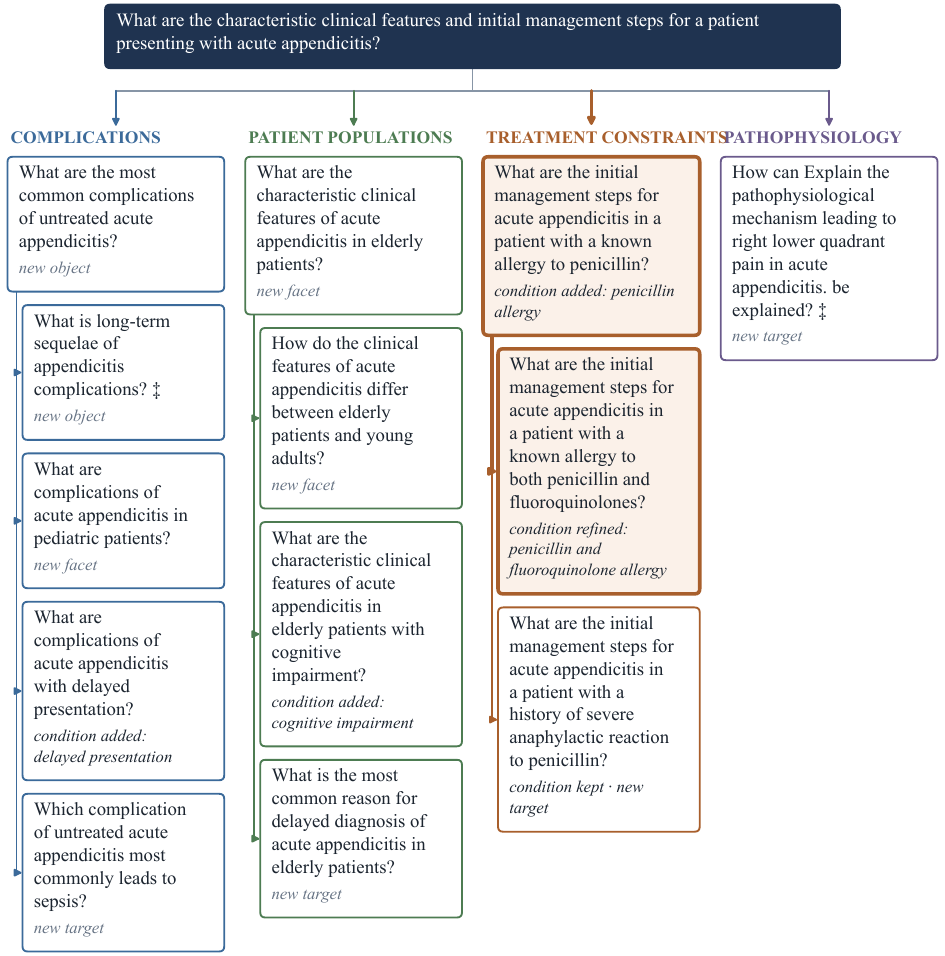}
\caption{
Complete executed acquisition tree for an acute
appendicitis root in Full MapSteal.
All 14 executed nodes are shown with their original
questions and recorded parent--child links.
The selected nodes displayed in the main-text
qualitative example are part of this tree.
}
\label{fig:appendicitis_complete}
\vspace{-10pt}
\end{figure*}

Figure~\ref{fig:appendicitis_complete} shows the
complete executed acquisition tree for the acute
appendicitis root used in the qualitative example
in Figure~\ref{fig:qualitative_acquisition}.
The tree contains 14 executed nodes, including
the initial root and 13 recursively acquired questions.
The main-text figure displays 10 selected nodes
to highlight branching and context-specific
follow-up questions.

The complete tree illustrates additional acquisition
directions and preserves the original generated questions,
including occasional grammatical artifacts.
All parent--child relationships correspond to recorded
execution links.

\subsection{Qualitative Condition Transitions}
\label{app:condition_transitions}

\begin{figure*}[t]
\centering
\includegraphics[width=0.8\textwidth]
{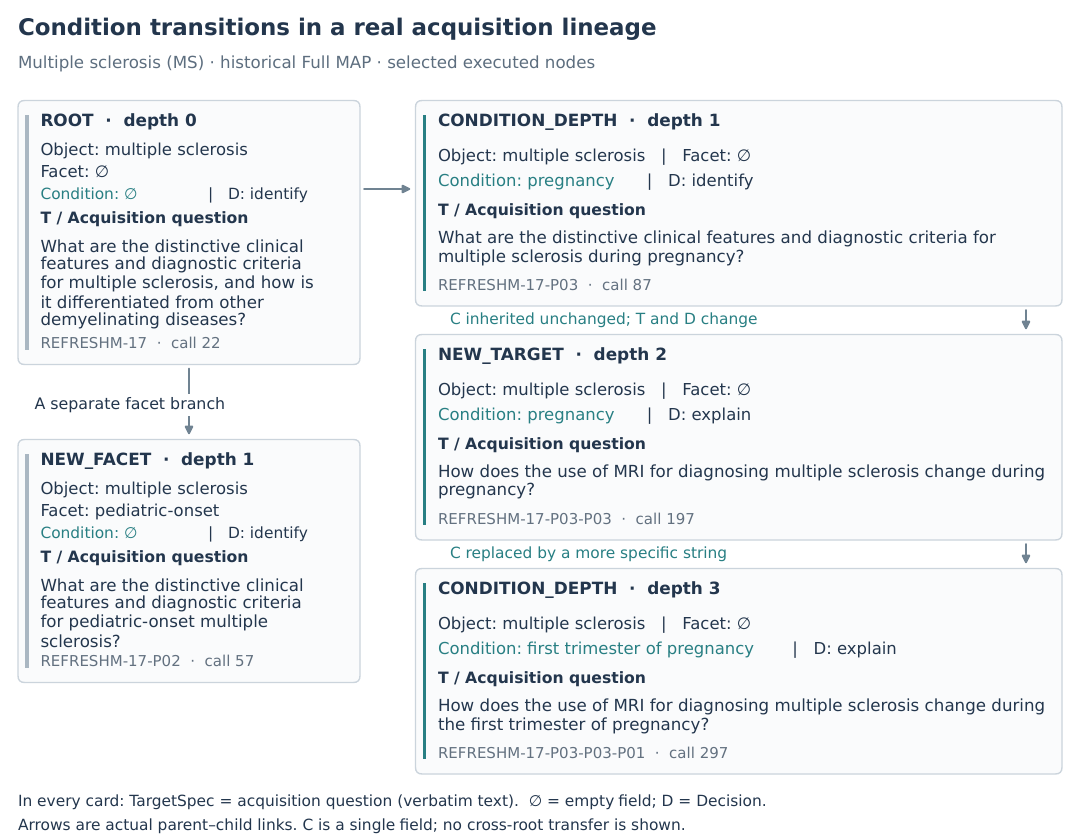}
\caption{
Condition transitions in a real acquisition lineage
rooted in multiple sclerosis.
CONDITION\_DEPTH introduces and later replaces
the condition, while NEW\_TARGET preserves the
existing condition.
A separate NEW\_FACET branch explores another facet.
All displayed nodes and edges are taken from the
executed acquisition trajectory.
}
\vspace{-10pt}
\label{fig:condition_transition}
\end{figure*}

We illustrate condition-aware recursive acquisition
using a real lineage rooted in multiple sclerosis,
as shown in Figure~\ref{fig:condition_transition}.
The CONDITION\_DEPTH operator first introduces
the condition \textit{pregnancy}.
A subsequent NEW\_TARGET expansion preserves
this condition while changing the acquisition
target toward MRI-based diagnosis.
A later CONDITION\_DEPTH expansion replaces
the condition with \textit{first trimester
of pregnancy}.
A separate NEW\_FACET branch explores
pediatric-onset multiple sclerosis.
These transitions illustrate how conditions
are introduced, inherited, and modified along
recorded acquisition lineages.

\section{Additional Cross-Domain Results}
\label{app:cross_domain_limits}

We examine whether recursive knowledge acquisition
improves downstream Student performance across
additional domains.
We distinguish the ability to expand acquisition
targets from the utility of the resulting supervision,
since generating more questions does not necessarily
improve downstream accuracy.

\subsection{Cross-Domain Results}

We measure the incremental effect of acquisition
as $\Delta_{\mathrm{acq}}$, defined as the accuracy
difference between the expanded training corpus
and its matched root-only or seed-only baseline,
in percentage points (pp).

We primarily evaluate Qwen3.5-4B and additionally
examine 2B and 9B Students where available.
The domains use different roots, evaluation panels,
and, in some cases, adaptation configurations.
Therefore, comparisons across domains are descriptive
rather than controlled estimates of domain effects.
Unless otherwise noted, three training seeds share
one Teacher acquisition trajectory.
Reported confidence intervals reflect
evaluation-item uncertainty conditional on
the observed runs, rather than variation across
independent Teacher acquisitions.

\begin{table*}[t]
\centering
\caption{
Cross-domain results with Qwen3.5-4B.
$\Delta_{\mathrm{acq}}$ measures the accuracy
difference over the matched root-only baseline,
except for Medicine, which uses Seed40-only.
Acquisition statistics are computed over the
first 460 successful executions, including
initial roots.
Confidence intervals are reported in
percentage points.
}
\label{tab:cross_domain_limits}

\small
\setlength{\tabcolsep}{5pt}

\begin{tabular}{lrrcccll}
\toprule
\textbf{Domain}
& \textbf{Roots}
& \textbf{Admit/call}
& \textbf{Novelty}
& $\boldsymbol{\Delta_{\mathrm{acq}}}$
& \textbf{95\% CI}
& \textbf{Seeds}
& \textbf{Panel} \\
\midrule

Medicine
& 35 & 2.57 & 0.179
& $+1.59$ & $[+0.37,+2.82]$
& 3 & Medical3 \\

CS
& 46 & 1.97 & 0.250
& $-0.57$ & --
& 3 & MMLU-CS \\

Law
& 48 & 1.72 & 0.257
& $-2.35$ & $[-3.57,-1.14]$
& 3 & LawMacro \\

Biology
& 42 & 2.39 & 0.213
& $+0.35$ & --
& 1 & MMLU-Bio \\

Business
& 64 & -- & --
& $-0.20$ & $[-2.31,+1.89]$
& 1 & MMLU-Business \\

Chemistry
& 48 & 2.18 & 0.257
& $+0.99$ & $[-0.55,+2.47]$
& 3 & Broad3 \\

Cybersecurity
& 40 & 2.31 & 0.268
& $+0.88$ & $[+0.27,+1.49]$
& 3 & CyberMacro \\

\bottomrule
\end{tabular}

\vspace{3pt}

\begin{minipage}{0.98\textwidth}
\footnotesize
Medicine compares B500 against Seed40-only,
whereas the other domains compare B500 against
answered roots.
Medical3 averages MedMCQA, MedQA, and
MedXpertQA-Text.
Chemistry Broad3 averages MMLU Chemistry,
MMLU-Pro Chemistry, and GPQA Main Chemistry.
CyberMacro averages CyberMetric and SecEval.
Training configurations are not identical
across all domains.
Novelty is measured sequentially over executed
acquisition questions.
Business has no corresponding harmonized
acquisition statistics.
A dash indicates that an interval or statistic
was not reported.
Confidence intervals do not account for
independent Teacher acquisition trajectories.
\end{minipage}

\end{table*}

\paragraph{Domain-Dependent Acquisition Gains.}

Table~\ref{tab:cross_domain_limits} shows that
the effect of recursive acquisition varies
across domains.
Medicine achieves a $+1.59$ pp improvement
over Seed40-only supervision, while
Cybersecurity improves by $+0.88$ pp
over its matched root-only baseline.
Chemistry exhibits a smaller positive difference
of $+0.99$ pp, although its confidence interval
includes zero.

In contrast, CS and Business show small
negative differences, and Law exhibits the
largest observed decline at $-2.35$ pp.
Biology shows a marginal positive difference
in the reported 4B evaluation.
These results indicate that recursive acquisition
can generate additional supervision without
consistently improving downstream performance.

\subsection{Results Across Student Scales}

We further examine acquisition gains across
2B, 4B, and 9B Students in Chemistry and
Cybersecurity.
Table~\ref{tab:cross_domain_scaling} reports
the incremental improvements over matched
root-only baselines.

\begin{table*}[t]
\centering
\caption{
Incremental acquisition gains across Student
scales (B500 minus root-only, pp).
Primary evaluation uses Broad3 for Chemistry
and CyberMacro for Cybersecurity.
Independent2 is an additional Chemistry
evaluation averaging ChemBench MCQ and
SuperGPQA Chemistry.
}
\label{tab:cross_domain_scaling}

\small
\setlength{\tabcolsep}{6pt}

\begin{tabular}{llcc}
\toprule
\textbf{Domain}
& \textbf{Student}
& \textbf{Primary $\Delta$ [95\% CI]}
& \textbf{Independent2 $\Delta$ [95\% CI]} \\
\midrule

Chemistry
& 2B
& $-0.89\;[-2.62,+0.81]$
& $-0.36\;[-1.19,+0.47]$ \\

& 4B
& $+0.99\;[-0.55,+2.47]$
& $-0.36\;[-1.20,+0.47]$ \\

& 9B
& $+1.79\;[+0.04,+3.49]$
& $-0.77\;[-1.69,+0.12]$ \\

\midrule

Cybersecurity
& 2B
& $+1.51\;[+0.78,+2.24]$
& -- \\

& 4B
& $+0.88\;[+0.27,+1.49]$
& -- \\

& 9B
& $+1.13\;[+0.56,+1.70]$
& -- \\

\bottomrule
\end{tabular}

\vspace{2pt}

\begin{minipage}{0.98\textwidth}
\footnotesize
All entries compare matched training runs
at the same Student scale under the
corresponding adaptation configuration.
Confidence intervals use paired
evaluation-item bootstrap resampling.
The three training seeds share one
acquisition trajectory within each domain
and Student scale.
\end{minipage}

\end{table*}

\paragraph{Scale and Benchmark Dependence.}

Cybersecurity shows positive acquisition gains
across all three Student scales, with improvements
of $+1.51$, $+0.88$, and $+1.13$ pp for
2B, 4B, and 9B, respectively.
The gains do not increase monotonically
with model size.

Chemistry exhibits greater variation.
On Broad3, the acquisition difference changes
from $-0.89$ pp at 2B to $+0.99$ pp at 4B
and $+1.79$ pp at 9B.
Although the 9B confidence interval narrowly
excludes zero, this exploratory comparison
was not corrected for multiple testing.

On Independent2, the corresponding differences
are $-0.36$, $-0.36$, and $-0.77$ pp,
with all confidence intervals including zero.
Thus, improvements on Broad3 do not consistently
extend to other Chemistry evaluation panels.

\subsection{Additional Domain Observations}

\paragraph{Chemistry.}

The Chemistry acquisition used 48 exam-style
roots drawn from OpenStax, MIT, and Harvey
materials.
Its B500 corpus contained 48 answered roots
and 452 recursive QA pairs, with no records
removed during quality control.
Over the first 460 executed states, acquisition
admitted 2.18 child targets per call, reached
a mean depth of 1.84, and achieved sequential
novelty of 0.257.

Despite continued recursive expansion,
the downstream effect depends on the
evaluation panel and Student scale.
The positive Broad3 result at 9B does not
extend to Independent2, where all three
Student scales show near-zero or negative
differences.
These observations illustrate that successful
acquisition expansion does not necessarily
produce supervision that transfers across
related benchmarks.

\paragraph{Law, Computer Science, and Biology.}
Law exhibits a $-2.35$ pp difference relative
to its root-only baseline despite continued
recursive expansion.
The harmonized trajectory admitted 1.72 child
targets per call and reached depth eight.
Without independent verification of acquired
QA quality, the cause of the decline remains
uncertain.
In Computer Science, Root46 improved MMLU-CS
by $+5.57$ pp over the pristine Student,
whereas expanding supervision to B500 produced
a $-0.57$ pp difference relative to Root46.
This suggests that additional recursive
supervision did not improve the evaluated
Student beyond the root-only baseline.
Biology shows similarly mixed evidence.
An extended B2000 trajectory achieved only
a $+0.56$ pp difference over Root42,
with a confidence interval including zero.
On a separate broader evaluation panel,
the B500--Root42 difference was $-2.21$ pp
(95\% CI $[-4.21,-0.33]$; one training seed).
Because the evaluations use different panels,
these results should not be treated as repeated
estimates of the same acquisition effect.

\subsection{Interpretation}

The cross-domain results distinguish
recursive acquisition activity from the
downstream usefulness of acquired supervision.
Admission rates, recursion depth, and semantic
novelty do not consistently predict performance
changes.
For example, Medicine shows relatively low
sequential novelty but a positive acquisition
gain, whereas Law maintains recursive expansion
without a corresponding downstream improvement.
Domains differ in root construction, evaluation
panels, and adaptation configurations, and most
use a single Teacher acquisition trajectory.
Overall, structured recursive acquisition expands
domain-specific knowledge targets, and its
downstream benefit depends on the acquired
supervision and the evaluation setting.


\section{Baseline and Resource Details}
\label{app:baseline_details}

\subsection{Run-to-Run Variability of EvoKD}
\label{app:evokd_variance}
\begin{table}[t]
\centering
\caption{
Individual EvoKD runs with Qwen3.5-2B.
Results are accuracy (\%); the final row reports
mean $\pm$ sample SD over three independent runs.
}
\label{tab:evokd_2b_runs}

\small
\setlength{\tabcolsep}{5pt}

\begin{tabular}{lccc}
\toprule
& \multicolumn{3}{c}{\textbf{Medicine}} \\
\cmidrule(lr){2-4}
\textbf{Run}
& \textbf{MedMCQA}
& \textbf{MedQA}
& \textbf{MedXpert} \\
\midrule
Seed 42 & 31.20 & 31.42 & 10.69 \\
Seed 43 & 44.80 & 45.56 & 12.69 \\
Seed 44 & 45.60 & 46.58 & 13.80 \\
\midrule
& $40.53\pm8.09$
& $41.19\pm8.47$
& $12.39\pm1.57$ \\
\midrule
& \multicolumn{3}{c}{\textbf{Cybersecurity}} \\
\cmidrule(lr){2-4}
\textbf{Run}
& \textbf{CyberMetric}
& \textbf{SecEval}
& \textbf{CTI} \\
\midrule
Seed 42 & 77.83 & 76.59 & 61.68 \\
Seed 43 & 65.75 & 67.06 & 42.91 \\
Seed 44 & 69.43 & 64.83 & 53.59 \\
\midrule

& $71.00\pm6.19$
& $69.49\pm6.25$
& $52.73\pm9.42$ \\
\bottomrule
\end{tabular}
\end{table}

Table~\ref{tab:evokd_2b_runs} reports the
individual EvoKD results using the
Qwen3.5-2B Student on the medical and
cybersecurity benchmarks.

Unlike MapSteal, which uses three training
seeds on one fixed acquisition trajectory,
EvoKD uses three independent acquisition
and online adaptation runs.
Its reported variation therefore combines
changes in acquisition and Student training.

The 2B results exhibit substantial run-to-run
variation, particularly on MedMCQA, MedQA,
and CTI-Bench.
In Medicine, seed 42 produces lower accuracy
than the other two runs on MedMCQA and MedQA.
In Cybersecurity, seed 43 shows the largest
decline on CyberMetric and CTI-Bench,
whereas seed 44 obtains the lowest SecEval
accuracy.
All three recorded runs are included in the
reported results, without excluding
low-performing trajectories.
Since acquisition and online adaptation
vary jointly, these observations do not
isolate the source of the variability.

\subsection{Teacher Cost Accounting}
\label{app:teacher_cost}

Table~\ref{tab:teacher_cost} summarizes
recorded Teacher token usage and reconstructed
acquisition costs for the three methods.
All methods operate under a nominal budget
of approximately 500 QA supervision instances,
although their acquisition procedures and
training corpus sizes differ.

These statistics describe the recorded
API usage rather than a controlled
cost-efficiency comparison.

\begin{table}[tbph]
\vspace{-10pt}

\centering
\small

\caption{
Recorded Teacher token usage and estimated
acquisition costs under the nominal B500
supervision budget.
Token counts are reported in thousands (K).
Costs are reconstructed from archived usage
records rather than verified invoices.
Some EvoKD records have incomplete
Teacher-call accounting.
}
\label{tab:teacher_cost}

\resizebox{\columnwidth}{!}{
\begin{tabular}{llrrrr}
\toprule

\textbf{Domain}
& \textbf{Method}
& \textbf{QA Rows}
& \textbf{Input (K)}
& \textbf{Output (K)}
& \textbf{Cost (\$)} \\

\midrule

\multirow{3}{*}{Medicine}
& MapSteal
& 495 & 1278.6 & 318.3 & 5.10 \\

& EvoKD
& 504
& $132.3\pm24.8$
& $54.8\pm8.3$
& $0.70\pm0.12$ \\

& QEDKS
& 500
& $439.8\pm6.4$
& $53.9\pm1.5$
& $1.31\pm0.02$ \\

\midrule

\multirow{3}{*}{Cybersecurity}
& MapSteal
& 500 & 1268.9 & 294.1 & 4.89 \\

& EvoKD
& 504
& $119.6\pm2.4$
& $50.3\pm0.3$
& $0.64\pm0.01$ \\

& QEDKS
& 500
& $241.6\pm0.2$
& $55.1\pm0.2$
& $0.92\pm0.00$ \\

\bottomrule
\end{tabular}
}
\vspace{-10pt}
\end{table}

Costs are calculated using the recorded pricing
of \$2 and \$8 per million input and output
tokens, respectively.
For Medicine, the MapSteal total includes
\$0.2602 for the 40 previously acquired
cold-start QA pairs and excludes
root-construction costs, which accounts for
the difference from the acquisition-only
cost in Table~\ref{tab:root_count_statistics}.

MapSteal reports one acquisition trajectory,
whereas QEDKS and EvoKD report means and
sample standard deviations across recorded
acquisition runs.
Four archived EvoKD run records contain
fewer logged calls than the planned schedule;
their token counts and costs are therefore
lower bounds.
MapSteal's per-record token usage is higher
because each request carries the structured
state and the direction registry. The methods
differ in acquisition procedure and training
schedule, so the costs are reported as recorded
usage.


\section{Preliminary Failure Attribution Diagnostic}
\label{app:failure_guided_analysis}

Before developing MapSteal, we examined whether
incorrect Student predictions consistently indicate
transferable knowledge deficits.
In a preliminary Medicine diagnostic with Qwen3.5-4B,
we evaluated error persistence under independent
question reformulation.

\begin{table}[tbph]
\centering
\caption{
Error persistence under question reformulation
in a preliminary Medicine diagnostic.
}
\label{tab:failure_persistence}
\small
\setlength{\tabcolsep}{5pt}
\begin{tabular}{lcc}
\toprule
\textbf{Metric} & \textbf{Count} & \textbf{Rate} \\
\midrule
Initially incorrect & 195 & -- \\
Excluded cases & 16 & -- \\
Evaluated errors & 179 & 100.0\% \\
\midrule
Errors persisted & 50 & 27.9\% \\
Correct after reformulation & 129 & 72.1\% \\
\bottomrule
\end{tabular}
\end{table}

Of the 195 initially incorrect examples,
16 ambiguous or unsuitable cases were excluded.
Among the remaining 179 examples, only 50
(27.9\%) remained incorrect after reformulation,
while 129 (72.1\%) were answered correctly.
These results suggest that individual Student
errors may be sensitive to question formulation
rather than reflecting stable knowledge deficits.
This preliminary diagnostic motivates
Student-independent acquisition but does not
establish that Student-aware methods are ineffective.

\end{document}